\documentclass[10pt,twocolumn,letterpaper]{article}

\usepackage{cvpr}              

\makeatletter
\@namedef{ver@everyshi.sty}{}
\makeatother

\usepackage{graphicx}
\usepackage{amsmath}
\usepackage{amssymb}
\usepackage{booktabs}
\usepackage[printonlyused]{acronym}

\usepackage{lipsum}
\usepackage[utf8x]{inputenc}
\usepackage{pgfplots}
\usepackage[export]{adjustbox}
\pgfplotsset{compat=1.17}
\usepackage{siunitx}

\usepackage[pagebackref,breaklinks,colorlinks]{hyperref}

\usepackage[capitalize]{cleveref}
\crefname{section}{Sec.}{Secs.}
\Crefname{section}{Section}{Sections}
\Crefname{table}{Table}{Tables}
\crefname{table}{Tab.}{Tabs.}

\newcommand{\binomial}{binomial}
\newcommand{\gaussian}{Gaussian}
\newcommand{\icdar}{ICDAR}
\newcommand{\lowpass}{low-pass}
\newcommand{\sheq}{shift equivariance}
\newcommand{\mobilenet}{MobileNetV2}

\def\cvprPaperID{8} 
\def\confName{ECCV Workshop on Text in Everything}
\def\confYear{2022}

\begin{document}

\title{Shift Variance in Scene Text Detection}

\author{Markus Glitzner,
        Jan-Hendrik Neudeck,
        Philipp Härtinger\\
MVTec Software GmbH\\
www.mvtec.com\\
{\tt\small \{markus.glitzner, jan-hendrik.neudeck, philipp.haertinger\}@mvtec.com}
}
\maketitle

\begin{abstract}
   Theory of convolutional neural networks suggests the property of shift equivariance, i.e., that a shifted input causes an equally shifted output. In practice, however, this is not always the case. This poses a great problem for scene text detection for which a consistent spatial response is crucial, irrespective of the position of the text in the scene.  
   
   Using a simple synthetic experiment, we demonstrate the inherent shift variance of a state-of-the-art fully convolutional text detector. Furthermore, using the same experimental setting, we show how small architectural changes can lead to an improved shift equivariance and less variation of the detector output. We validate the synthetic results using a real-world training schedule on the text detection network. To quantify the amount of shift variability, we propose a metric based on well-established text detection benchmarks.
   
   While the proposed architectural changes are not able to fully recover shift equivariance, adding smoothing filters can substantially improve shift consistency on common text datasets. Considering the potentially large impact of small shifts, we propose to extend the commonly used text detection metrics by the metric described in this work, in order to be able to quantify the consistency of text detectors. 

\end{abstract}

\begin{figure}[t]
  \begin{subfigure}{0.495\columnwidth}
  \centering
  \includegraphics[width=1\textwidth]{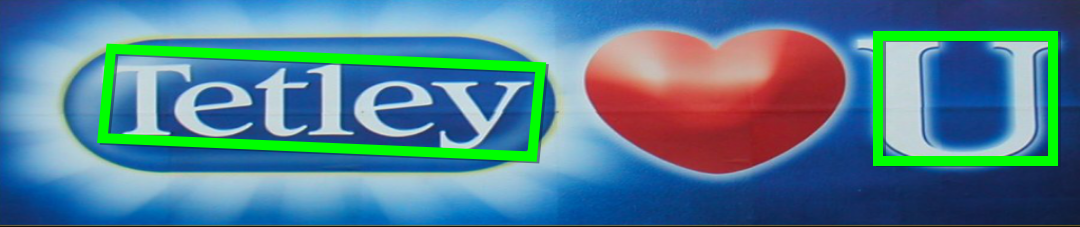}
  \caption{$\text{HMean}=100\%$} 
  \end{subfigure}
  \hfill
  \begin{subfigure}{0.495\columnwidth}
  \centering
  \includegraphics[width=1\textwidth]{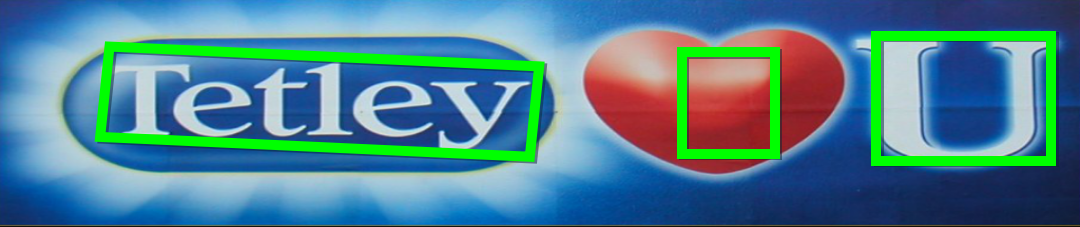}
  \caption{$\text{HMean}=80\%$} 
  \end{subfigure} 
  \begin{subfigure}{0.495\columnwidth} 
  \includegraphics[width=\textwidth]{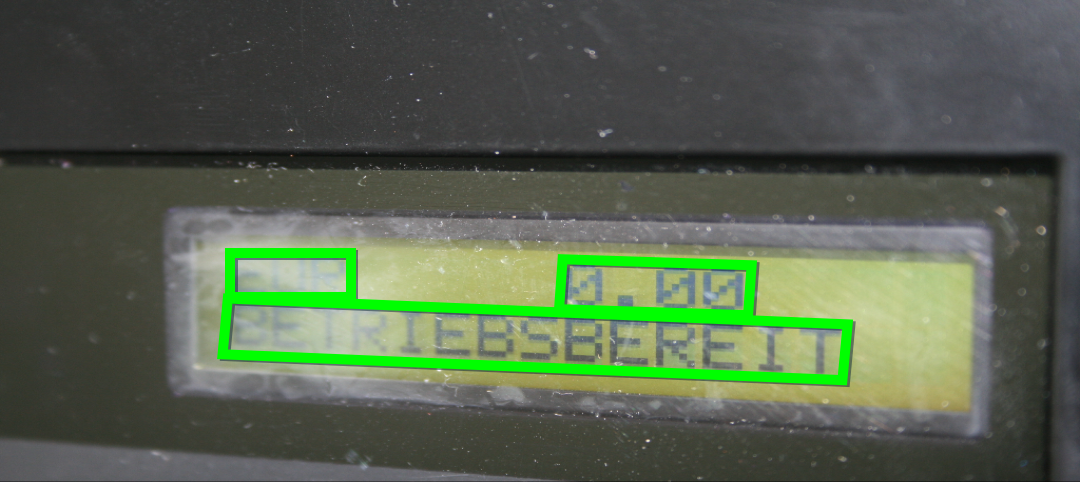}
  \caption{$\text{HMean}=100\%$} 
  \end{subfigure}  
  \hfill 
  \begin{subfigure}{0.495\columnwidth} 
  \includegraphics[width=\textwidth]{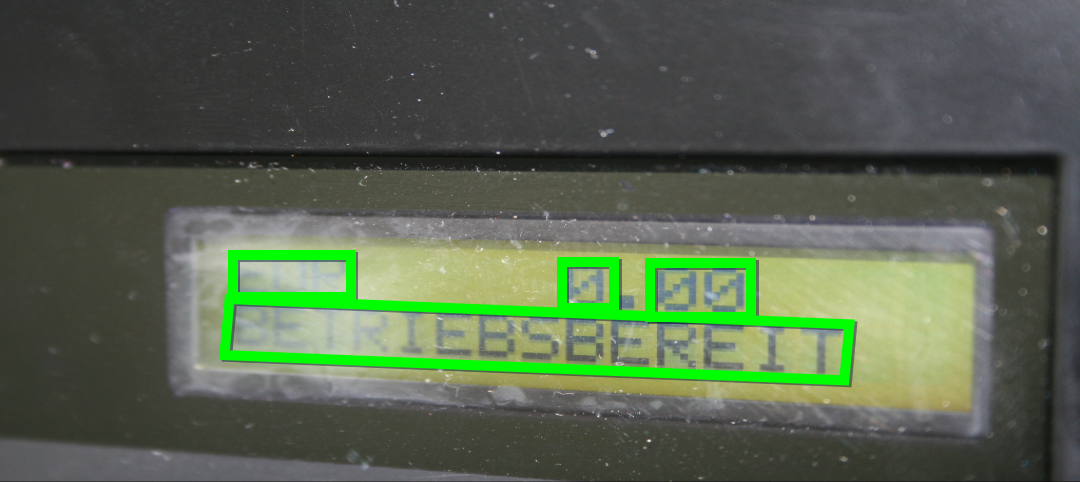}
  \caption{$\text{HMean}=57.1\%$} 
  \end{subfigure}
  \caption{Examples from the \acl{IC13} \cite{karatzas2013icdar} test set. Shifting the input image by a single pixel can produce strongly different text detection results. For this experiment, we used the state-of-the-art text detector CRAFT \cite{baek2019craft}. 
  Common errors are false positive detections (top) and split detections (bottom).}
  \label{fig:clova_icdar_2013_shift_variance}
\end{figure}
\section{Introduction}
\label{sec:intro}

The majority of current deep learning methods in image processing are based on discrete convolutions operating on discrete feature domains. While in digital signal processing \cite{oppenheim99} the effects of filtering spatially limited domains with discrete filters are well understood, their impact on deep CNNs is widely neglected. As suggested by the notion of a single kernel striding over an input feature map, filtering images with cascaded convolutions is assumed to be equivariant with respect to shifts of the input image. Therefore, a shift in the input domain should result in an equally shifted output feature domain.

Recently, there have been investigations on the impact of signal processing effects on deep learning classification. It was shown how signal aliasing \cite{zhang2019making, chaman2021truly}, kernel padding \cite{alsallakh2021mind} and kernel size \cite{alsallakh2021debugging} impose significant shift variance on classification consistency. Also, preliminary reports discuss how object detection performance can be affected by the spatial position of objects in the input image \cite{manfredi2020equivariance, alsallakh2021debugging}.

For \ac{STD}, however, the phenomenon was not yet described.
In this work, we investigate how shift variance affects text detection models, as illustrated in \cref{fig:clova_icdar_2013_shift_variance}.
Using publicly available scene text datasets \cite{gupta2016synthtext, karatzas2013icdar, karatzas2015icdar} we train a state-of-the-art segmentation-based text detector \cite{baek2019craft}.
To simulate text occurring at variable positions, we evaluate multiple shifted versions of each image in the scene text datasets.
We show that for commonly used metrics \cite{karatzas2015icdar}, the output of the text detector varies significantly even for small input shifts of only one or two pixels.
To grade the quality of a text detection model with respect to {\sheq}, we introduce a metric to quantify the degree of shift variance for a given text detection model.
Moreover, we present simple architectural changes to reduce the degree of shift variance for an exemplary \ac{STD} architecture, namely the type of downsampling in the encoder part of the network.


\begin{figure*}[!t]
    \subfloat[input location $(x_i, y_i)=(128, 128)$]{%
        \centering
        \includegraphics[width=0.33\textwidth]{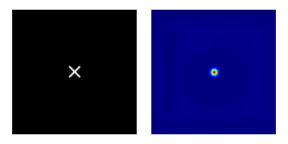}
        \label{toy_example_128}}
    \hfill
    \subfloat[input location $(x_i, y_i)=(129, 128)$]{%
      \centering
      \includegraphics[width=0.33\textwidth]{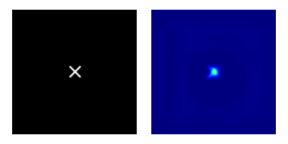}
      \label{toy_example_129}}
    \hfill
    \subfloat[Maximum response in output map.]{%
      \centering
      \includegraphics[max width=0.28\textwidth]{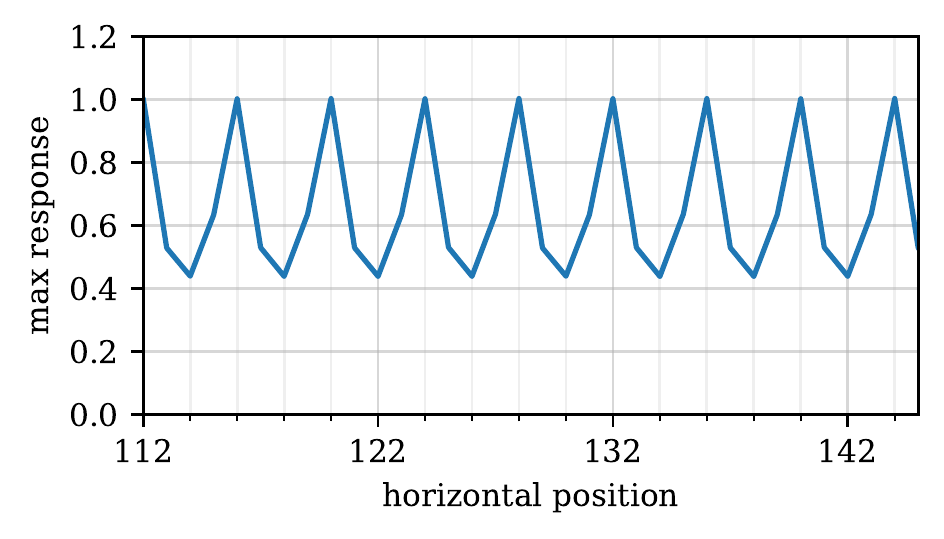}
      \label{toy_example_max_reponse}
      }
    \caption{Translation of the input object by one pixel $\left(\text{\protect\subref{toy_example_128}} \rightarrow \text{\protect\subref{toy_example_129}}\right)$: A large response variation is observed in the score map. Training data for this experiment contains objects at every fourth input position. Assuming {\sheq} of fully convolutional networks, the output is expected to result in a uniform response. However, measuring the maximum response \protect\subref{toy_example_max_reponse} reveals significant shift variance.}    \label{fig:toy_example_default}
\end{figure*}

\section{Methods}
\label{sec:methods}

\subsection{Scene text detection model architecture}
For our experiments we use the CRAFT \cite{baek2019craft} text detection model, which predicts character and link score maps.
Targets for character and link score maps are generated as described in \cite{baek2020character}.
During post-processing, these score maps are used to create an oriented word box for each word.
\\
Deviating from the reference model, we substitute the VGG-16 \cite{Simonyan15} backbone with a {\mobilenet} \cite{sandler2018mobilenetv2} for increased computational efficiency.
Following the official model design, five stride $2$ convolutions are used for spatial downsampling in the encoder path of the CRAFT architecture.

\subsection{Demonstrating shift variance for \ac{STD}}
\label{subsec:toy_experiments}
To demonstrate that the shift variance in \ac{STD} is strongly affected by the applied downsampling strategy and less by the training data and optimization settings, we design a minimal synthetic experiment.
For this, we generate a dataset containing images where the symbol '$\times$' is drawn in white color on a black background.
The detection model is trained to generate an isotropic {\gaussian} response $\in [0,1]$ with its peak at the character center.
Using this simple setting, we have full control over the spatial distribution of the character within the dataset and can ensure that all possible locations are seen during training.

Assuming {\sheq} for fully convolutional architectures \cite{long2015fcn}, training with samples that cover only a subset of the input canvas should result in a uniform detection response, independent of the text position.
We experiment with training on samples that contain characters at every fourth input pixel.
During inference, we shift the character pixel-wise in horizontal direction.
We evaluate the consistency of the model by measuring the dependence of the signal response on the input position.
For this, we record the maximum response in the output map for each pixel-wise shifted input image.

As shown in \cref{fig:toy_example_default}, the position dependency becomes clearly visible in the oscillating network output. 
Here, the maximum response varies between $0.4$ and $1.0$ with a periodicity of four pixels -- the same periodicity as the input data during training.
This proves that the assumption of inherent {\sheq} does not apply for this fully convolutional model architecture.
In practice, the inconsistent model response can potentially result in decreased detection performance or lead to unpredictable behaviour, i.e., shift inconsistency, as it was shown in \cref{fig:clova_icdar_2013_shift_variance}.

\subsection{Quantifying shift consistency}
\label{subsec:shift_metrics}
Our goal is to measure the consistency of the model, i.e., the variance of the \ac{STD} performance subject to the position of the text in the input image.
To that end, we generate pixel-wise shifted samples from the \ac{IC13} and \ac{IC15} test set.
We resize the samples in an aspect-ratio preserving manner to
$\left(H+2r, W+2r\right)$
with $H$ and $W$ being the model input height and width, respectively, and
$r\in\mathbb{N}$ being the maximum symmetric shift range.
Then, we generate $2r+1$ crops of size $\left(H,W\right)$ from the sample along the horizontal image dimension.
We exclude samples for which a bounding box would exceed the image limits in any of the crops to ensure that all text is fully visible for all shifts.
We therefore evaluate on $157$ of $233$ \ac{IC13} and $269$ of $500$ \ac{IC15} samples.
All validations are conducted on equally sized images with a resolution of $1024\times1024$ and $2048\times2048$ for \ac{IC13} and \ac{IC15}, respectively.

\begin{figure*}[!t]
    \subfloat[strided convolution]{%
      \centering
      \includegraphics[max width=0.32\textwidth]{figures/aa_comparison_stride_4_default_eval_size_256_obj_type_x_iter_6800.pdf}
      \label{toy_strided_conv}}
    \hfill
    \subfloat[{\binomial} smoothing filter]{%
      \centering
      \includegraphics[width=0.32\textwidth]{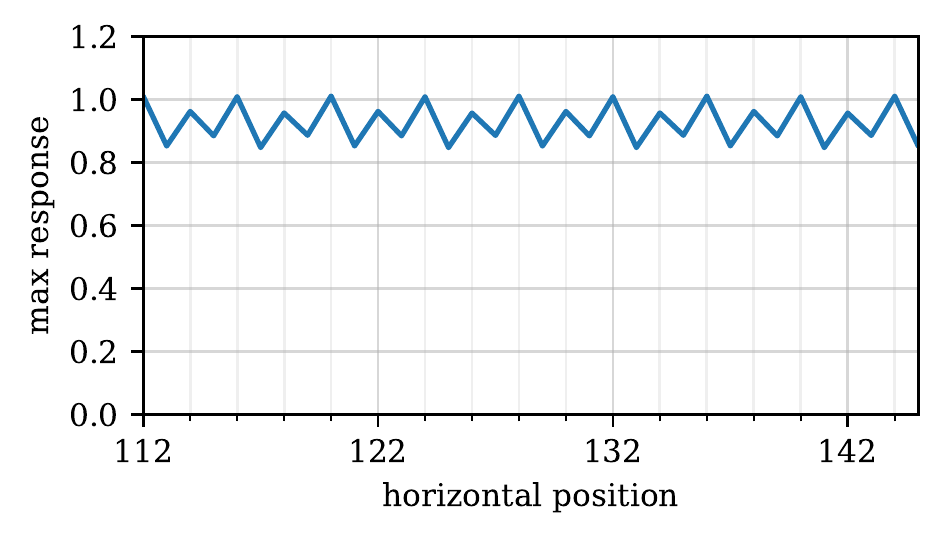}
      \label{toy_aa}}
    \hfill
    \subfloat[average pooling]{%
      \centering
      \includegraphics[max width=0.32\textwidth]{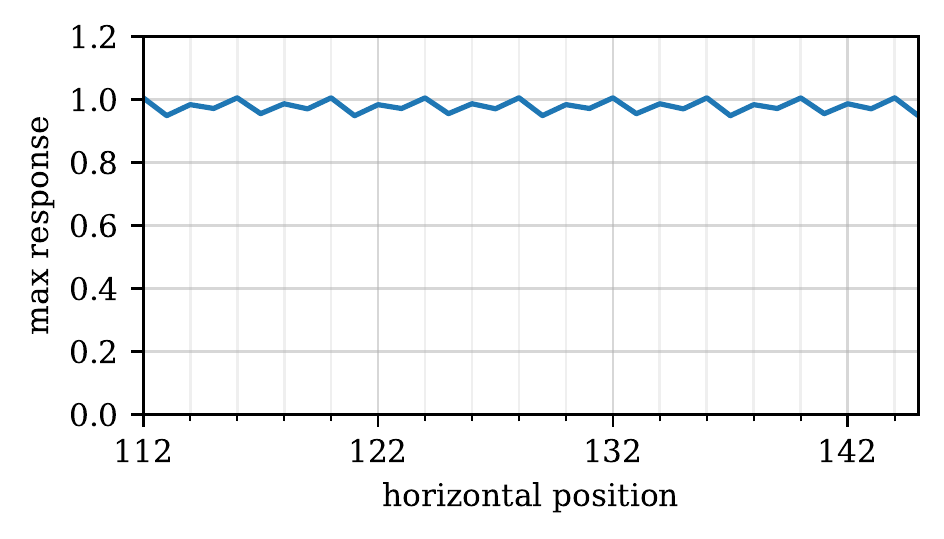}
      \label{toy_avg_pooling}}
    \caption{Evaluation of shift variance for different downsampling strategies using the experimental settings of \cref{subsec:toy_experiments}. The baseline {\mobilenet} backbone architecture implements strided convolution \protect\subref{toy_strided_conv}. Here, the model output is profoundly position dependent. Adding $k=5\times5$ {\binomial} filter kernels as {\lowpass} filter before every downsampling stage reduces the effect \protect\subref{toy_aa}. Instead of strided convolution, we further experiment with strided $3\times3$ average pooling layers which are able to suppress the majority of the position dependency \protect\subref{toy_avg_pooling}.}
    \label{fig:striding methods}
\end{figure*}

For assessing the text detection results, we use the well-established \ac{IC15} evaluation protocol for text localization \cite{karatzas2015icdar} using a $50\%$ Intersection-over-Union threshold.
In order to quantify the shift consistency of the model, we evaluate the spread of the harmonic mean (HMean) of Precision and Recall.
Following \cite{manfredi2020equivariance}, we define $\Delta\text{HMean}$ as the difference between best and worst HMean for all possible shifts in a range $\left[-r,r\right]$:
\begin{equation}
  \Delta\text{HMean}(r) = \text{HMean}_{\text{max}}(r) - \text{HMean}_{\text{min}}(r)
\end{equation}
with $\text{HMean}_{\text{min}}$ and $\text{HMean}_{\text{max}}$ being the minimum and maximum HMean, respectively, within the shift range $r$.\\
The backbone architecture halves the input resolution five times in succession. 
Hence, to move the output by a whole pixel on the lowest layer, the input must be shifted by $32$ pixels.
Therefore, a natural choice for the maximum shift range in our experiments is $r=16$. 

\subsection{Robust downsampling methods}
\label{sec:robust_downsampling_methods}
Using $3\times3$ convolutions with stride $2$ for spatial downsampling, as done in {\mobilenet}, causes an undersampling of the input signal that can provoke aliasing effects.
To counteract these effects, we introduce {\lowpass} filters before every downsampling stage in order to precondition the input feature map before applying downsampling \cite{zhang2019making}.
We implement a $5\times5$ {\binomial} filter, approximating a {\gaussian} smoothing filter.\\
Another way to counteract aliasing is to replace the stride $2$ convolution by a $3\times3$ average pooling layer with stride $2$, followed by a $3\times3$ convolutional layer with stride $1$.
Similar to the {\binomial} filters, we use strided average pooling as a simple {\lowpass} operation for suppressing aliasing effects during downsampling.\\
\Cref{fig:striding methods} shows the impact of signal conditioning on the response consistency for shifted input signals using the experimental setting described in \cref{subsec:toy_experiments}.

\subsection{Experiments}
\label{subsec:icdar_experiments}
We verified the presence of shift variance for text detection in a toy example in \cref{subsec:toy_experiments}. Additionally, in \cref{sec:robust_downsampling_methods} we showed how shift consistency can be recovered at least partially using low-pass filters for signal preconditioning.
Motivated by these results, we want to measure the influence of shift variance on real-world problems.
For this purpose, we train the text detection model following the same setting as \cite{baek2019craft}, i.e., using SynthText \cite{gupta2016synthtext} as pretraining dataset and refine on the training splits of \ac{IC13} and \ac{IC15}. We use translation, rotation, scaling and color augmentation throughout.
The goal of the experiments is to quantify and reduce the degree of shift variance in state-of-the-art networks by adapting the downsampling layers in the backbone.
As we do not focus on beating the current best text detection methods in this work, no extensive hyperparameter search is conducted.
\\
We evaluate the trained models on the shifted test splits of \ac{IC13} and \ac{IC15} as described in \cref{subsec:shift_metrics}.

\section{Results}
\label{sec:results}

\begin{figure*}[!htbb]
     \subfloat[Shift variance evaluation for \ac{IC13} test set. Lower absolute values are better.]{%
      \centering
      \includegraphics[width=0.99\textwidth]{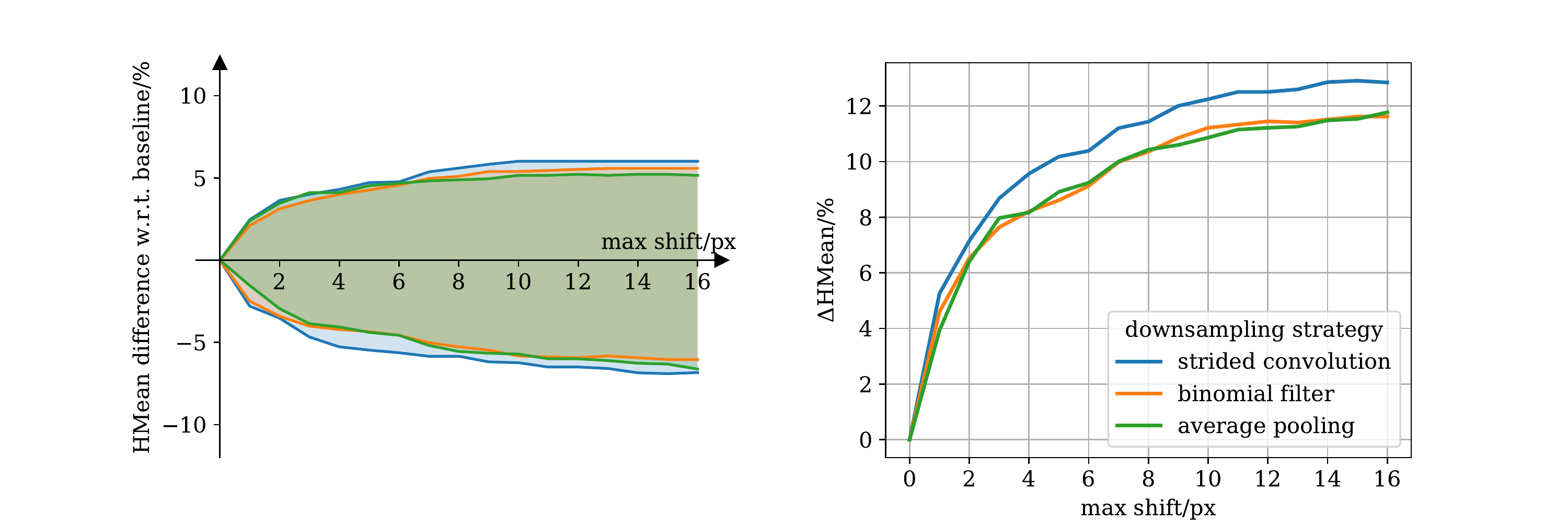}
      \label{fig:icdar_13_spread}}
    \hfill
    \subfloat[Shift variance evaluation for \ac{IC15} test set. Lower absolute values are better.]{%
      \centering
      \includegraphics[width=0.99\textwidth]{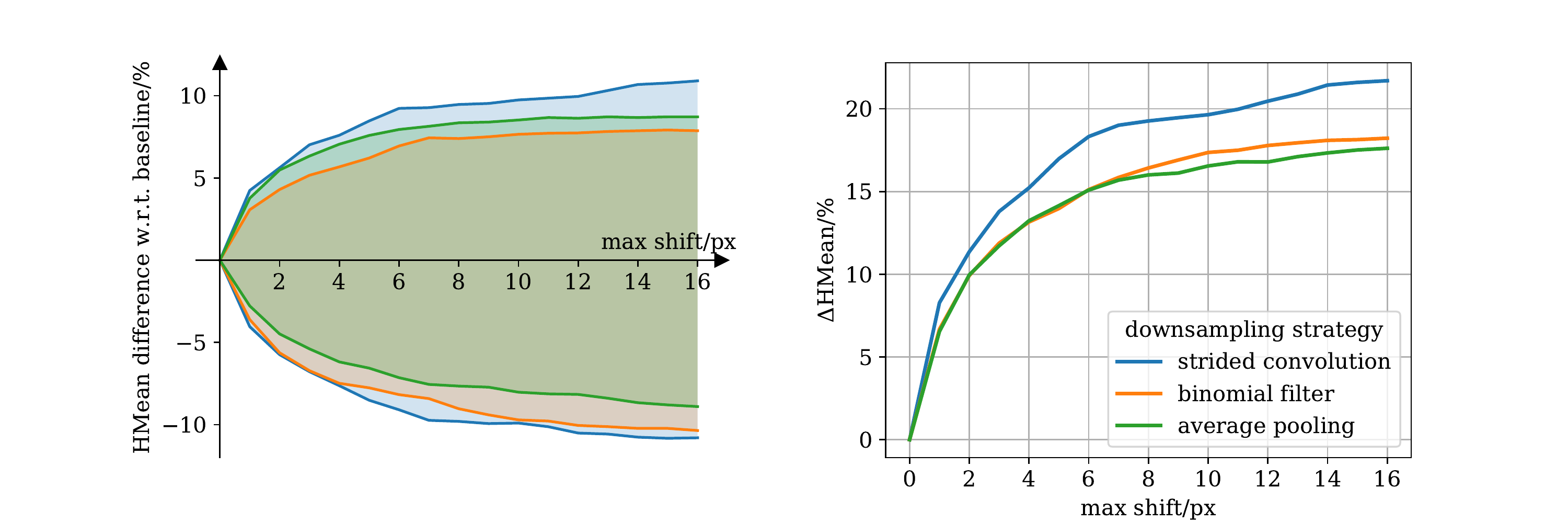}
      \label{fig:icdar_15_spread}}
    \caption{Shift variance evaluation for \ac{IC13} and \ac{IC15} test set. Left, the minimum and maximum HMean deviation from the baseline (shift $= 0$) dependent on the maximum absolute shift is plotted. On the right, $\Delta\text{HMean}$ for each downsampling strategy is provided. The horizontal axis indicates respectively the maximum absolute shift, i.e., how far a sample is maximally shifted in both the positive and negative direction. For both datasets, replacing the strided convolutions in the backbone by low-pass filters reduces the shift variance.}
    \label{fig:spreads}
\end{figure*}

\Cref{fig:icdar_13_spread} shows the spread of HMean with greater shift ranges.
The default strided convolution based downsampling causes an alarming $\Delta\text{HMean}$ of $\SI{12.91}{\percent}$.
This means that the best and worst case scenario can make nearly $\SI{13}{\percent}$ difference, depending only on the position of the text in the input image.
Using the described methods causes the maximum $\Delta\text{HMean}$ to drop to $\SI{11.62}{\percent}$ or $\SI{11.77}{\percent}$ for \binomial{} filter and average pooling, respectively.

For the more complex \ac{IC15} dataset, the improvement is even higher (see \cref{fig:icdar_15_spread}). 
Here, the strided convolution based downsampling causes significant shift variance, resulting in a maximum $\Delta\text{HMean}$ of $\SI{21.69}{\percent}$.
Preconditioning the feature maps with \binomial{} filters reduces the maximum performance difference to $\SI{18.22}{\percent}$. 
Strided average pooling results in a $\Delta\text{HMean}$ of $\SI{17.61}{\percent}$.

Besides the reduction of shift variance, the overall detection performance also benefits from the slight architectural changes (see \cref{tab:hmean_icdar}).
For both \ac{IC13} and \ac{IC15} datasets the improved {\sheq} naturally correlates with an enhanced HMean score. 

\begin{table}[htbp]
\caption{Evaluation of \ac{IC15} metric on the \ac{IC13} and \ac{IC15} test sets with different downsampling strategies implemented in the model backbone. Augmenting the default strided convolution with smoothing filters improves the overall model performance.}
\centering
\begin{tabular}{l|l|l}
 downsampling strategy & {\icdar} 13 & {\icdar} 15 \\
 \hline
 strided convolution &  \SI{81.88}{\percent} & \SI{72.24}{\percent} \\
 \binomial{} filter & \SI{82.07}{\percent} & \SI{73.17}{\percent} \\
 average pooling & \SI{82.01}{\percent} & \SI{73.58}{\percent} \\
\end{tabular}
\label{tab:hmean_icdar}
\end{table}

\section{Discussion}
\label{sec:discussion}
We described simple architectural changes to reduce the effects of shift variance in \ac{STD}.
Compared to {\gaussian} low-pass filtering, strided average pooling is a lightweight and natural choice for conditioning and downsampling input features in one step. Contrary, conditioning with {\gaussian} filtering produces an intermediate result at full resolution and therefore occupies more memory \cite{zhang2019making}. 
Both signal conditioning methods could reduce the expected uncertainty of a detector's accuracy and therefore improve shift consistency. This was especially true for the \ac{IC15} dataset, where each sample contains more but relatively small text.
Previously it was shown that strong data augmentation like random cropping, translation, and rotation can reduce the effect of shift variance \cite{azulay2018deep,hendrycks2019augmix}.
In contrast to our synthetic experiments in \cref{subsec:toy_experiments}, these augmentations are commonly applied for large scale scene text detection training, consequently also in our trainings.
The small difference in baseline performances over the different architectures can be caused by this learnt type shift equivariance which is likely to only apply to test data from the same distribution as the training data \cite{azulay2018deep}.

In this work, we used samples from publicly available scene text datasets that were synthetically shifted. 
This does not necessarily represent the full complexity of real-world scenarios in which text occurs at various positions in the input image.
For further investigations, we propose to create a dedicated dataset which covers text that is pixel- and subpixel-wise shifted. 
This would enable the field to not only quantify the performance of text detectors based on static images but also quantify the consistency with respect to sample shifts, as the proposed metric in \cref{subsec:shift_metrics} suggests.

{\small
\bibliographystyle{ieee_fullname}
\bibliography{egbib}
}

\acrodef{STD}{Scene Text Detection}
\acrodef{IB}{Inverted Bottlenecks}
\acrodef{IC13}{\icdar{} 2013}
\acrodef{IC15}{\icdar{} 2015}

\end{document}